\documentclass[sigconf]{acmart}
\usepackage{algorithm}
\usepackage{moreverb,url}
\usepackage{mathpartir}
\usepackage{xspace}
\usepackage[frozencache,cachedir=.]{minted2}
\usepackage{simplebnf}

\newcommand{\egglog}{\texttt {egglog}\xspace}
\newcommand{\egg}{\texttt {egg}\xspace}
\newcommand{\babble}{\textsc {babble}\xspace}

\DeclareMathAlphabet{\mathcal}{OMS}{cmsy}{m}{n}

\newcommand{\C}[1]{\mathsf{#1}} 
\newcommand{\V}[1]{\mathit{#1}} 

\renewcommand\footnotetextcopyrightpermission[1]{} 
\acmConference[Preprint]{}{2026}{X}
\AtBeginDocument{%
  }

\begin{document}

\title{Library learning with e-graphs on jazz harmony}

\author{Zeng Ren}
\affiliation{%
  \institution{EPFL}
  \city{Lausanne}
  \state{Vaud}
  \country{Switzerland}
}
\email{zeng.ren@epfl.ch}

\author{Maddy Bowers}
\affiliation{%
  \institution{MIT}
  \city{Cambridge}
  \state{Massachusetts}
  \country{USA}
}
\email{mlbowers@mit.edu}

\author{Xinyi Guan}
\affiliation{%
  \institution{EPFL}
  \city{Lausanne}
  \state{Vaud}
  \country{Switzerland}
}
\email{xinyi.guan@epfl.ch}

\author{Martin Rohrmeier}
\affiliation{%
  \institution{EPFL}
  \city{Lausanne}
  \state{Vaud}
  \country{Switzerland}
}
\email{martin.rohrmeier@epfl.ch}


\begin{abstract}
Humans can acquire a highly structured intuitive understanding of musical patterns, yet these patterns often require multiple iterations of reflection and re-listening to internalize fully. To capture such an internalization process, we present a computational model for the learning of jazz harmonic patterns based on library learning. Given a corpus of harmonic progressions, our model searches over a space of programs composed of primitive harmonic relations in order to discover concise generative explanations of the corpus. The model first enumerates possible programs for each piece, and then jointly learns a library of harmonic patterns and refactored programs. To efficiently navigate the vast joint space of programs and libraries, we integrate deductive parsing with library learning on e-graphs. We explore how well our model captures aspects of human musical pattern learning by evaluating the intuitiveness of both programs and libraries, as well as similarities to human-written harmonic derivations.
\end{abstract}



\maketitle
\section{Introduction}

Musical patterns are central to how music is composed, perceived, and analyzed. In music theory, patterns such as motives, schemata, cadential formulas, harmonic progressions and rhythmic cells provide reusable units of description that support both analytic interpretation and compositional practice. In music information retrieval (MIR), pattern discovery likewise serves as a foundational problem \cite{janssen2013finding}, underpinning tasks such as motivic/thematic search \cite{hsiao2023bps,hall2021model,ross2012detecting,cambouropoulos2006musical,lartillot2007motivic}, structural segmentation \cite{nieto2020audio,de2013structural,sargent2011regularity,paulus2010state,paulus2006music}, similarity \cite{chakrabarty2022approach,pearce2017compression,boot2016evaluating,volk2012melodic,ziv2007themes}, and style characterization \cite{zhang2022influence,de2007pattern,van2005musical,cope2002recombinant}. More broadly, pattern learning is a hallmark of human musical cognition: listeners form expectations, recognize returns, and construct higher-level structure by relating present material to prior context. For these reasons, computational models of musical pattern are not only practically useful, but also theoretically informative as hypotheses about the representations and inductive biases that make musical understanding possible.


This paper focuses on \emph{musical pattern discovery} (MPD): the problem of identifying recurrent musical regularities directly from data. A central challenge in MPD is to find patterns that are (1) abstract enough to subsume non-exact repetition and (2) analytically meaningful to support interpretation and structural understanding. These requirements frequently come into conflict. As a pattern becomes more abstract, it tends to match more passages; without appropriate inductive bias, the space of matches quickly fills with false positives and the method returns patterns that are trivial or musically uninformative. Conversely, highly specific patterns can be precise but brittle, failing to capture the family-resemblance structure characteristic of musical variation. To address these challenges systematically, we ask: \emph{what makes a musical pattern interesting in general?} 
We argue that a broadly applicable answer should not be stated only in terms of counts or surface similarity. Instead, it should connect pattern quality to the role that abstractions play in building compact, reusable explanations of musical structure.


\subsection{Our perspective and approach}\label{sec: learning paradigm}

Humans can acquire highly structured musical patterns from musical surface, but such structure is rarely learned in a single listening. Understanding typically emerges through iterative cycles of hypothesis and revision: listeners (and analysts) propose candidate abstractions, revisit the material, and refine those abstractions in light of how well they account for the musical surface. On this view, pattern discovery is inseparable from the broader process of constructing an interpretation of a piece or corpus.

To make this idea precise, we adopt a view from computational cognitive science in which understanding is modeled as building structured \emph{programs} that explain observations \cite{chater2013programs,tenenbaum2011grow,lake2015human}. These programs are generative and compositional: they can produce many surface forms from a small set of parts. 
This allows us to state concrete objectives. We want explanations that fit the data, but we also want them to be simple---for example, short programs or low description length. The program representation also supports explicit operations that mirror analysis, such as factoring out repeated subcomputations and introducing reusable library components. From this perspective, a ``good'' musical pattern is an abstraction that is useful in many explanations. It earns its place in the model because it can be reused and because it reduces the overall description length of the corpus. Patterns are therefore not defined by frequency alone; they are defined by utility for compression and explanation.
This perspective motivates the guiding question of this work: can we learn jazz harmonic patterns from chord progressions alone, in an unsupervised way, by treating pattern discovery as the discovery of reusable program fragments?

The central computational question of this study is that given only the observed musical surface (e.g., chord progression), can we infer jointly (1) a compact generative program (derivation) for each piece and (2) a set of shared abstractions that are reusable across pieces? In other words, can we simultaneously construct structured explanations of individual works and discover the higher-level building blocks that make those explanations economical?

Traditionally, these two problems correspond to separate research traditions. The first problem (inferring a structured derivation from surface form) corresponds to \emph{parsing}, which typically assumes a fixed grammar and seeks the best derivation of a given surface sequence under that grammar. The second problem, which involves discovering reusable abstractions from a collection of programs corresponds to \emph{library learning}. In program synthesis, one begins with a corpus of programs and compresses them by introducing shared subroutines that reduce overall description length. Parsing presupposes a stable set of primitives; library learning presupposes a corpus of programs as input. In our setting, however, this separation is artificial. If the goal is to learn harmonic patterns directly from surface data, then the space of derivations depends on the available abstractions, while the usefulness of any abstraction depends on how it participates in derivations of specific pieces. The two problems are therefore mutually dependent.

\begin{figure*}[t]
    \centering
    \includegraphics[width=\textwidth]{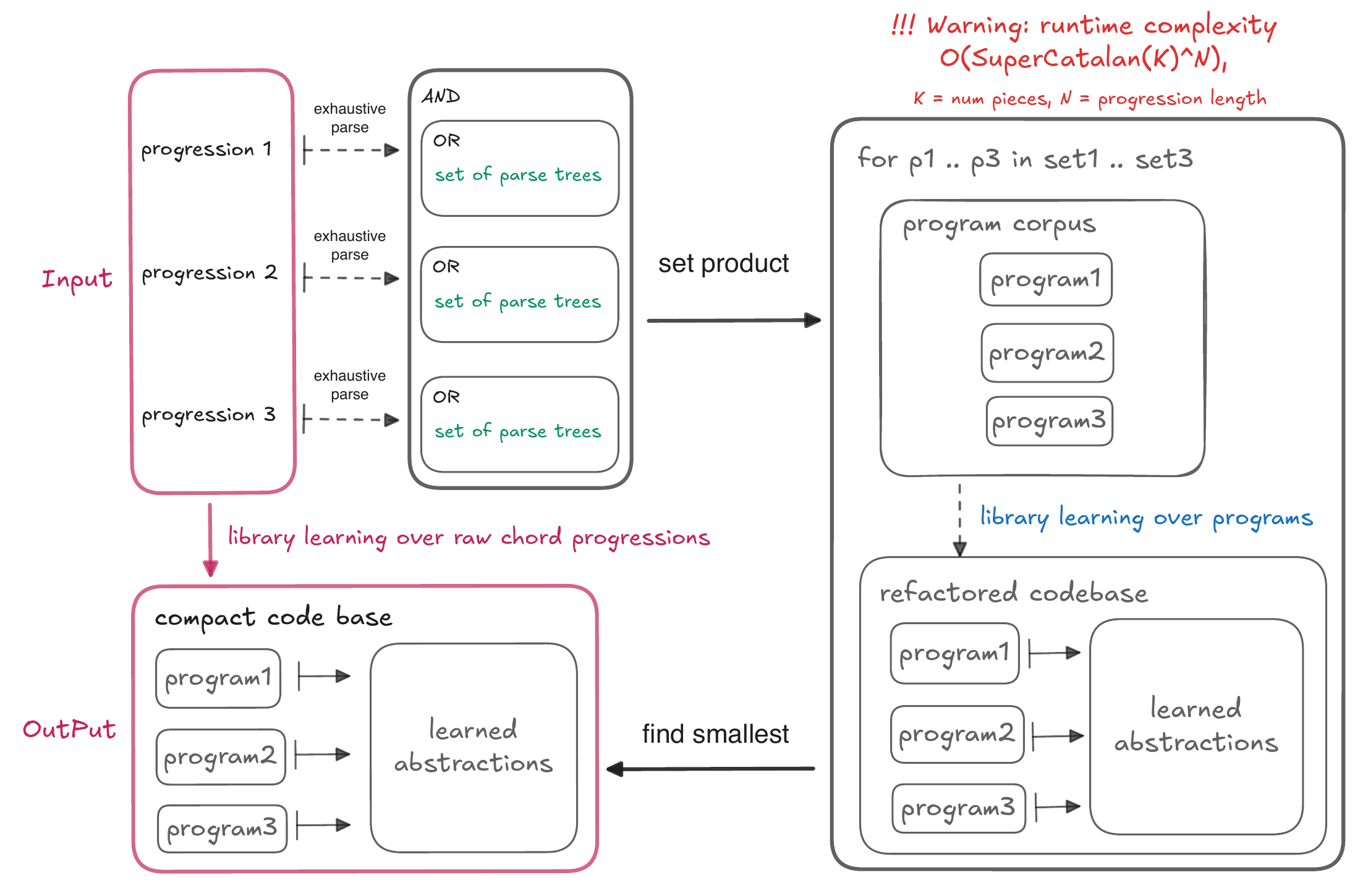}
    \caption{One naive adaptation of library learning (highlighted in blue) to account for chord progression as input (instead of programs). This adaptation contains multiple sources of combinatorial explosion.}
    \label{fig: naiveAdaptationOfLibraryLearning}
\end{figure*}

A naive solution would proceed sequentially (see Figure~\ref{fig: naiveAdaptationOfLibraryLearning}). One might first exhaustively enumerate all parse trees for each chord progression, treating each derivation as a candidate program. Library learning could then be applied to this corpus of programs to identify shared substructures and introduce new abstractions. Although conceptually straightforward, this pipeline is computationally intractable. For binary-branching ambiguous CFGs, the number of parses of a string of length $k$ grows as the Catalan number $C_{k-1}$; for a corpus of $n$ independent pieces, the space of possible program corpora scales accordingly $C_{k-1}^N$.
Therefore, exhaustively enumerating derivations before abstraction leads to a runtime explosion.


This study proposes an unsupervised approach to jazz harmonic pattern discovery that integrate deductive parsing with library learning on e-graphs. E-graphs are compact data structures designed to represent and manipulate large families of equivalent programs via non-destructive term rewriting \cite{zhang2023better,cao2023babble}. By maintaining many candidate derivations in a shared representation, they allow abstraction and parsing decisions to interact without explicitly enumerating the full search space.
We evaluate the system along two complementary dimensions. Quantitatively, we measure \emph{compression rate} to assess how well the learned abstractions reduce the corpus's overall description length. Qualitatively, we examine musical plausibility of the learned patterns and the way they are used to explain specific pieces.

\section{Library learning on chord progressions}

\subsection{Problem Setup}

The previous section motivated musical pattern discovery as the search for reusable program fragments that support compact explanations of musical structure. The goal is to infer, from a corpus of chord progressions, both (1) a structured derivation (i.e., program) for each progression and (2) a shared set of harmonic abstractions that make those derivations economical. These two goals are mutually dependent: derivations determine which structures recur, while abstractions reshape the space of possible derivations. 

The first phase concerns parsing, which aims to infer all the possible derivation directly from chord sequences. We model a harmonic analysis as a program in a domain-specific language (DSL). A program is a compositional derivation that generates a chord sequence by recursively combining primitive harmonic relations (see Section~\ref{sec: DSL}). Since each progression typically admits many valid derivations under an ambiguous context-free grammar (CFG), the space of all possible programs that evaluate to the surface chord progression of the entire corpus can grow combinatorially. Rather than committing prematurely to a single derivation for each piece, we retain this ambiguity and reason over families of derivations via e-graph.

The second phase concerns finding a \emph{compact code base} for the entire corpus. If each progression were explained independently using only primitive relations, the resulting programs would often repeat similar substructures across pieces. Such redundancy signals the presence of higher-level regularities. The aim of library learning is to factor out these recurring structures into reusable abstractions, so that the corpus can be described more economically. The result is analogous to refactoring a code base: repeated fragments are extracted into named functions, reducing duplication and clarifying shared structure. 

\subsection{Harmonic relations as DSL primitives}\label{sec: DSL}

\begin{figure}[ht]
     \centering
     \includegraphics[width=0.8\linewidth]{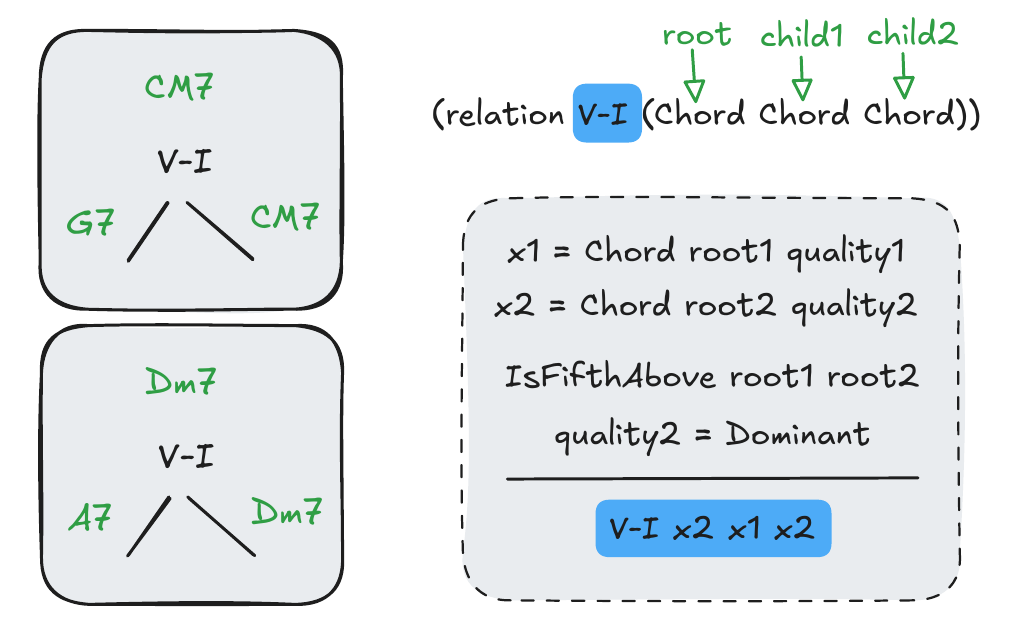}
     \caption{Representation of harmonic relations.}
     \label{fig:rep_harmonic_relations}
\end{figure}

We take an established grammar \cite{rohrmeier2020syntax} as our DSL for jazz harmony, but reinterpret its production rules in explicit relational terms. Rather than representing a harmonic analysis as a tree of chord symbols, we treat it as a composition of \emph{harmonic relations}. The chords themselves are structured objects, and the internal nodes of a derivation are logical predicates that connect them as illustrated in Figure~\ref{fig:rep_harmonic_relations}. For example, a \texttt{V-I} (or \texttt{Descending5th}) relation does not simply rewrite one chord symbol into another; it asserts that two chord instances satisfy intervallic and qualitative constraints (e.g., one root is a fifth above the other, one chord has dominant quality). This shift in emphasis is conceptually important. It makes explicit that harmonic structure lies in how chords stand in systematic relations to one another.
Formally, each primitive is encoded as a logical rules in \egglog \cite{zhang2023better}, which provides a logic programming interface over e-graphs (see Listing 1 in Appendix for the encoding of the \texttt{Descending5th} relation).



\subsection{Combing deductive parsing and library learning e-graphs}

\begin{figure*}[t]
    \centering
    \includegraphics[width=1\linewidth]{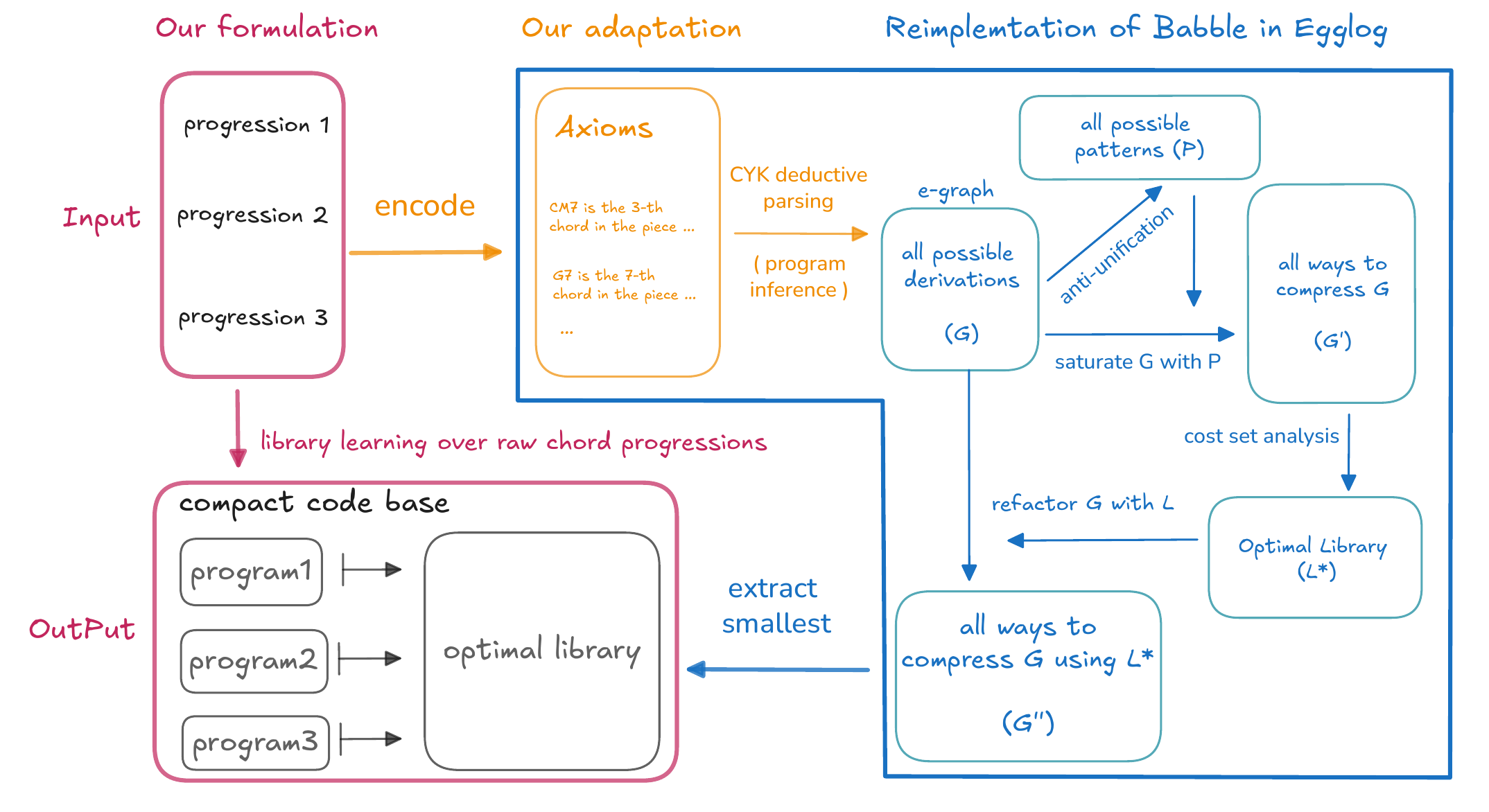}
    \caption{We mitigate the combinatorial explosion by adapting the \babble algorithm, which uses e-graphs to share common sub-expressions. We first do exhaustive parsing of all the pieces (each as one chord progression) under the e-graph structure. The rest follows the \babble algorithm except that all terms are Template program as opposed to lambda terms}
    \label{fig: ourAdaptationOfBabble}
\end{figure*}

Our model integrates deductive parsing \cite{shieber1995principles,pereira1983parsing} and library learning \cite{cao2023babble} into a single architecture. Figure~\ref{fig: ourAdaptationOfBabble} summarizes the full pipeline. Each chord progression is encoded as logical facts (axioms) and parsed using a CYK-style deductive procedure implemented in \egglog \cite{zhang2023better}. The resulting e-graph $\mathcal{G}$ compactly represents all derivations for all pieces, merging identical sub-derivations into equivalence classes. 
Candidate patterns ($\mathcal{P}$) are generated by anti-unification over e-classes. Anti-unification computes the least general generalization of two sub-derivations, resulting in a parameterized template that captures their shared structure. These candidates are added to the e-graph as rewrite rules, and we perform equality saturation \cite{tate2009equality,willsey2021egg}: the graph is expanded by exhaustively applying the rewrites until no new equivalent expressions can be derived. This produces a saturated graph $\mathcal{G}'$ that compactly represents not only all original derivations, but also all equivalent derivations obtainable using the candidate abstractions.

Given the saturated e-graph $\mathcal{G}'$, we evaluate candidate libraries using cost set analysis. Every e-node is associated with a cost set of the $(\text{cost}(\mathcal{L}), \text{cost}({\mathcal{D}\mid \mathcal{L}}))$. The first term, $\text{cost}(\mathcal{L})$, measures the complexity of introducing the abstractions in a library $\mathcal{L}$; the second term, $\text{cost}({\mathcal{D}\mid \mathcal{L}}))$, measures the cost of encoding the corpus $\mathcal{D}$ using those abstractions. Together, these terms instantiate a MDL objective: abstractions are favored only insofar as the savings they enable in encoding the data outweigh their own complexity. We select the library that minimizes total description length under the current search space:
\begin{equation}
\mathcal{L}^* = \arg \min_{\mathcal{L}\subseteq\mathcal{P}}(\text{cost}(\mathcal{L}) + \text{cost}({\mathcal{D}\mid \mathcal{L}}))
\end{equation}
Finally, we refactor the original e-graph $\mathcal{G}$ using the optimal library $\mathcal{L}^*$. The output of the model is therefore a compact code base: a learned library of harmonic patterns together with compressed derivations of each progression expressed in terms of that library.

\section{Implementation in \egglog}
Our implementation is built in \egglog \cite{zhang2023better}, a declarative extension of the equality saturation framework \egg. The original \egg system introduced equality saturation as an efficient method for equational reasoning over e-graphs, which allows large families of equivalent programs to be represented and optimized simultaneously \cite{willsey2021egg}. \egglog extends this line of work with bottom-up logic programming, integrating Datalog-style rule evaluation directly into the e-graph representation. 
This extension is critical for our purposes. The bottom-up evaluation strategy allows us to express CYK-style deductive parsing as logical rules whose consequences are accumulated in the e-graph. 
We can also explicitly store all the possible derivation trees in a compact way. As a result, all well-formed derivations permitted by the DSL are stored compactly as equivalence classes (e-classes) within a single shared structure.

\subsubsection{Parsing stage}

The parsing stage deduces and compactly stores all derivations of all pieces in the corpus under the given grammar. We implement this procedure declaratively in \egglog by running two inference rules (see Figure 1 in Appendix) to a fixed point (commonly referred to as saturation). This process instantiates a CYK-style dynamic program inside the e-graph, augmented to explicitly construct and merge derivation programs during inference.
Importantly, for each derived fact of the form \texttt{IsPhrase}, we identify all derivation programs that result in the same phrase as equivalent (denoted $\equiv$) to a canonical term \texttt{Der}. 
The resulting e-graph compactly merges equivalent derivations. After saturation, the e-graph therefore resembles a Shared Packed Parse Forest (SPPF)\cite{shieber1995principles}: it encodes a set of derivations that may be exponentially large under an ambiguous grammar.


A standard limitation of CYK-style parsing is that it generates all locally well-formed spans, including partial derivations that do not participate in any complete parse of the piece. Since our goal is to discover reusable structure from full derivations, we exclude such fragments through a subsequent filtering pass. This is implemented as a top-down propagation of root-connectedness.
The base case marks as root-connected the e-class corresponding to derivations that span the entire piece. Inductively, if a phrase spanning $(i,k)$ is root-connected and was derived by merging phrases spanning $(i,j)$ and $(j,k)$, then those child phrases (and their associated derivation e-classes) are also marked as root-connected. This backward propagation ensures that only derivations participating in at least one complete parse are retained for subsequent library learning.




\subsubsection{Anti-unification stage}


Once the e-graph has been populated with derivations, the next step is to identify candidate abstractions. Intuitively, we seek recurring structural patterns across derivation programs. For example, the expressions $f (3 + 42) \times g (42)$ and $f(x^2+7) \times g(7)$ share the common structure $f(X+Y)\times g(Y)$, where constants and subexpressions have been abstracted into variables.
A standard technique for discovering such shared structure is \emph{anti-unification} \cite{plotkin1970lattice,cerna2023anti}. Viewing programs as trees, anti-unification computes their least general generalization (i.e., the largest common rooted tree obtainable by replacing mismatched subtrees with variables). The result is a pattern that captures precisely the structure shared by both inputs. Recent work on \babble \cite{cao2023babble} extends the classical tree-based anti-unification \cite{plotkin1970lattice} to operate directly over an e-graph. Rather than comparing two individual trees, we compute anti-unifiers between equivalence classes of derivations. The computation is defined mutually recursively over e-nodes and e-classes and implemented via dynamic programming to avoid redundant work. Because \egglog combines equality saturation with bottom-up logic programming, the entire procedure can be expressed declaratively while remaining efficient.

A central challenge is controlling how anti-unifier information propagates through the e-graph. Naively propagating partial results can lead to redundant re-propagation when new function values are deduced. To avoid such cases, we interleave the computation of e-node and e-class anti-unifiers. The dependency is as follows: the anti-unifier of an e-node depends on the finalized anti-unifiers of its child e-classes; the finalized anti-unifier of an e-class depends on having collected all anti-unifier candidates produced by its constituent e-nodes. Thus, we alternate between (i) computing candidate anti-unifiers at the e-node level using finalized child information, and (ii) consolidating these into finalized e-class values once all node-level contributions are available.

The rules to formalize this process are provided in Figure 1 in Appendix. The base cases handle primitive relations: identical primitives anti-unify to themselves, while distinct primitives collapse to the identity pattern \texttt{Id}. The inductive cases handle composed templates. When two nodes share the same outer constructor and their children admit non-trivial anti-unifiers, these are composed to form a larger pattern (Success Inductive). If no informative generalization exists at the children, the result defaults to \texttt{Id} (Fail Inductive). The predicate \texttt{CanCoOccur} ensures that only derivations that can jointly participate in a complete parse are considered, preventing abstractions from unrelated fragments.

\subsubsection{Compression (refactoring) stage}

After the set of anti-unifiers ($\mathcal{P}$) has been computed for the corpus, we use these patterns to saturate the e-graph ($\mathcal{G}$) and perform rewrite on the saturated e-graph $\mathcal{G'}$. Consider again the pattern $f(X+Y)\times g(Y)$. This abstraction can rewrite 
$$f(3+42)\times g(42) \quad \text{as} \quad (\lambda x \ y. f(x+y)\times g(y)) \ 3 \ 42$$
In the Template program, the corresponding pattern is written
$$\times [ f  [+ ]
    , g  
    ]
$$
where square brackets denote structural composition (under the trivial meta-rules), rather than function application in the lambda-calculus sense. The brackets indicate that $\times$ composes two sub-templates, one headed by $f$, and the other by $g$, with a shared placeholder structure. The Template rewrite can also be expressed in a variable-free style:
$$\times  
    [ f  [+ [3,42]]
    , g  [42]
    ]  
    \quad \text{as} \quad
\times  
    [ f  [+ ]
    , g  
    ] \quad   \langle \ \_ \ \_ \ 1 \ \rangle \quad 
    [3,42]
$$
The angle-bracket operator specifies how arguments are routed from the input tuple into the template's open positions. In effect, refactoring is implemented as template rewrite over the e-graph, it replaces repeated structure with a higher-order combinator that reconstitutes the original program when supplied with the appropriate arguments.





\subsubsection{Cost set analysis of candidate libraries}

Taking all discovered patterns as library entries would generally be suboptimal. Many abstractions overlap or compete to explain the same structure, and each incurs a storage cost. The central problem is therefore one of trade-off: which combination of abstractions minimizes total description length when both library complexity and usage cost are taken into account? Exhaustively enumerating all subsets of candidate patterns would be computationally infeasible. Instead, \babble evaluates library choices compositionally, by propagating cost information through the e-graph.

The key idea is to associate each term in the e-graph (both e-nodes and e-classes) with a \emph{cost set}: a set of pairs $(\mathcal{L}, c)$, where $\mathcal{L}$ is a candidate subset of library entries and $c$ is the corresponding use cost. 
For primitive terms, which contain no substructure and require no abstractions, the cost set contains a single element $({\emptyset,1})$. For composite terms, cost sets are constructed bottom-up. The cost set of an e-node is obtained by combining the cost sets of its child e-classes: we take the cross product of the children's library–cost pairs and add the local contribution of the parent constructor (including any abstraction that may apply). This captures all consistent ways of explaining the parent given explanations of its parts. The cost set of an e-class is then defined as the union of the cost sets of its constituent e-nodes, since any equivalent derivation offers a valid explanation. As in the anti-unification stage, the computation is mutually recursive: node-level costs depend on finalized child classes, while class-level costs aggregate over alternative nodes.

This compositional structure fits naturally within \egglog's inference framework. Each rule specifies a local cost transformation, while the underlying relational engine efficiently performs the required joins (cross products) and unions. When multiple values are derived for the same term, they are merged via set union, accumulating alternative cost set pairs without explicitly enumerating global subsets of the library.

Since the number of possible library–cost combinations can grow rapidly, \babble prunes each e-class's cost set during bottom-up propagation using a beam-search heuristic, retaining only the most promising candidates under the MDL objective. As in earlier stages, we separate node-level accumulation from class-level finalization to ensure stable propagation and avoid redundant recomputation. The outcome is a compact representation of candidate libraries paired with their corpus-wide use costs.


\section{Results \& evaluation}

Applying the library learning procedure to the full corpus is not yet computationally feasible with the current algorithm. In particular, the \texttt{egglog}-based system is not sufficiently optimized to support large-scale joint inference over highly ambiguous pieces. We therefore present a proof-of-concept study on a small subset of three pieces learned jointly. This restricted setting allows us to examine the behavior of the model in detail and to inspect the musical plausibility of the learned abstractions and the derivations that use them.

The three pieces were chosen to have relatively low ambiguity (the number of possible derivations) under the base grammar, since ambiguity is the main source of combinatorial growth. The corpus consists of \emph{Red Clay} (length 13, 5 possible derivations), \emph{Valse Hot} (length 15, 6 derivations), and \emph{Sunny} (length 17, 31 derivations). For this experiment, we limit the maximum size of the learned library to 15 abstractions and use a beam width of 5 during search. 

\subsection{Quantitative results}

We first measure compression relative to derivations that use only the original grammar rules. Because the grammar is in Chomsky normal form, any derivation tree for a piece of length $n$ contains exactly $2n - 1$ nodes. This provides a fixed baseline description length. When a library is introduced, the total description length includes both (1) the size of the derivations expressed using library items and (2) the storage cost of the library itself. In the joint condition, the storage cost of the shared library is divided equally across the all pieces.

\begin{table}
\centering
\begin{tabular}{p{0.15\linewidth}  p{0.15\linewidth}   p{0.15\linewidth}   p{0.15\linewidth}  p{0.17\linewidth} }
\toprule
\textbf{Piece}                             & \raggedright{Derivation size w/o library} & \raggedright{Derivation size with library} & \raggedright{Storage cost per piece} & {Normalized CR} \\ \hline
\textbf{Red clay}  & 25  & \textbf{8} (9)                            & \textbf{10.33} (17)      & \textbf{1.36}  (0.96)       \\
\textbf{Valse hot} & 29                              & 12 (\textbf{8})                           & \textbf{10.33} (14)      & 1.29  (\textbf{1.32})          \\
\textbf{Sunny}     & 33                              & \textbf{7} (11)                            & \textbf{10.33} (16)     & \textbf{1.90} (1.22)           \\ \hline
\textbf{Total}     & 87                              & \textbf{27} (28)                          & \textbf{31} (47)                   & \textbf{1.5}  (1.16)                       \\ 
\bottomrule
\end{tabular}
\caption{Joint library learning on 3 pieces compared with local library learning on individual pieces (reported in parenthesis). Each piece's compression rate is calculated based on a equally shared the storage cost of the global library.}
\label{tab: library learning on 3 pieces}
\end{table}

Since our model outputs derivations that uses learned patterns, compression is measured relative to the size of derivation tree using only production rules from the grammar. As we use a simplified version of the jazz harmony grammar in Chomsky normal form, the size of any derivation tree for a piece with length $n$ is $2n-1$. This is a simple consequence that non-leafs nodes in a binary tree is one less than the leaf nodes.  
Table~\ref{tab: library learning on 3 pieces} compares joint and piece-wise library learning. For each entry, the main value reports the result of joint learning with a shared library, while the value in parentheses reports piece-wise learning, where each piece is processed independently without shared abstractions. In the joint condition, learned patterns can be reused across pieces, encouraging abstractions that capture recurring structure. In contrast, the piece-wise condition allows each piece to induce its own library, allowing more local and piece-specific patterns.

Joint learning substantially reduces the total derivation size, from 87 under the base grammar to 27 using the shared library. After accounting for storage cost, the normalized compression rate  (CR) is 1.5. In contrast, piece-wise learning yields a total derivation size of 28 but incurs a much larger overall storage cost (47 compared to 31), resulting in a lower compression rate of 1.16. Although individual runs can occasionally achieve slightly better compression for a single piece (e.g., in the case of ``Valse hot''), they do so by introducing more optimal abstractions that are specific to that piece and not reusable elsewhere.

\subsection{Qualitative results}

\begin{figure*}{th}
    \centering
    \includegraphics[width=\linewidth]{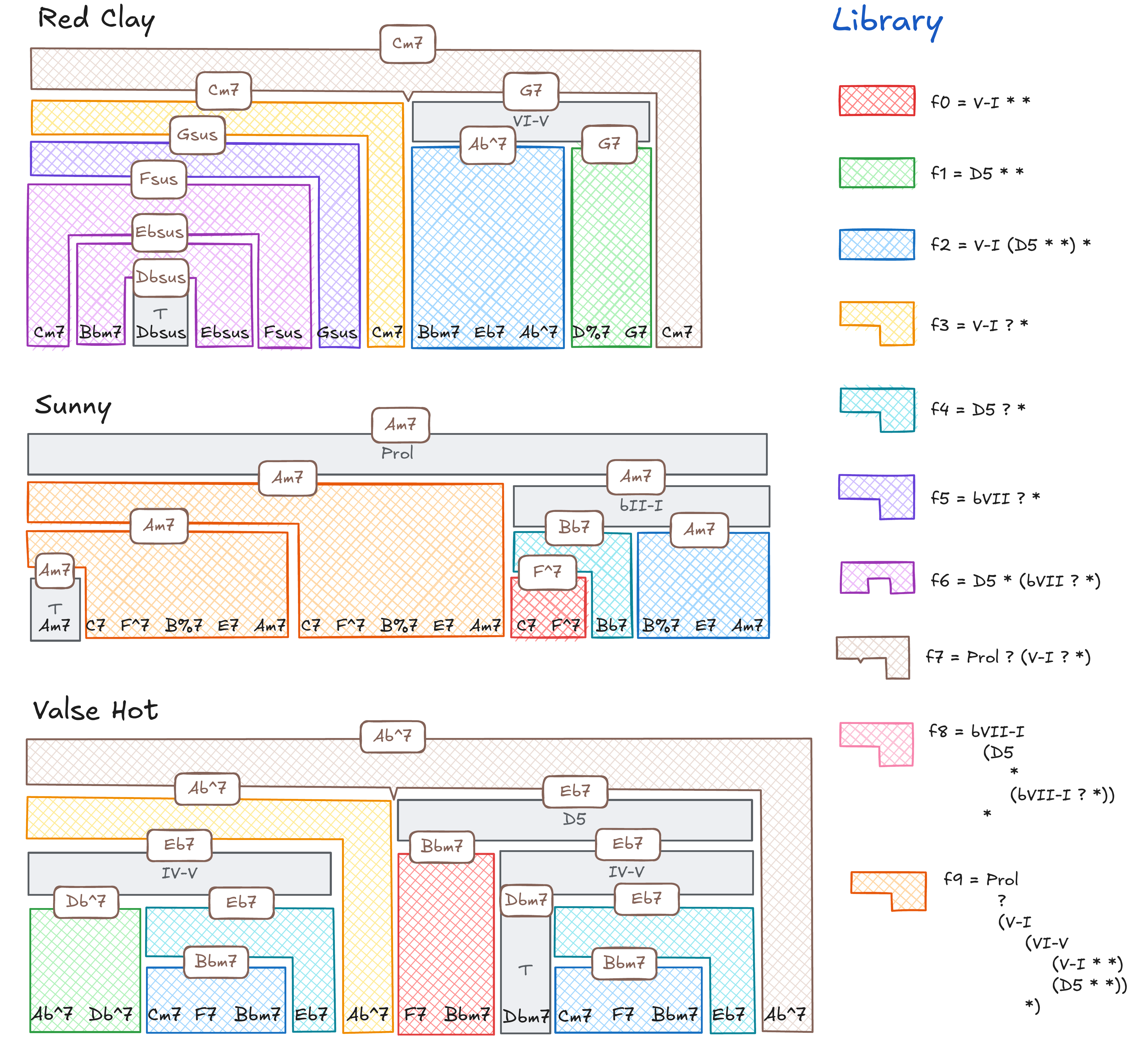}
    \caption{The model output derivations for the three pieces using the learned abstractions in the library. Question marks denote open arguments (fresh lambda variables) that can be filled by other patterns. Terminating rules are marked by ``*''. For readability, library functions are expanded into primitive productions, although in the learned library some functions are defined through composing other patterns (e.g., $f_2 = f_3 \circ f_1$, and $f_8 = f_5 \circ f_6$). }
    \label{fig: final derivations and library}
\end{figure*}

Figure~\ref{fig: final derivations and library} shows the learned abstractions and the derivations of the three pieces expressed in terms of them. The discovered patterns vary in size, arity, and number of terminals. Within the CFG framework, each abstraction can be understood as a new production rule that expands to a sequence of terminals and non-terminals. We visualize these abstractions as compositional blocks, making explicit both the surface material they consume and the structural positions they leave open.

Patterns that generate only terminals are shown as solid rectangles, which can be directly aligned with surface chord symbols. These include $f_0$, a concrete dominant-tonic succession; $f_1$, a descending fifth progression; and $f_2$, a concrete \texttt{ii-V-I} pattern with three terminals. Their presence in the learned library is expected: such progressions are highly frequent in jazz harmony, and therefore offer substantial compression when abstracted as reusable units. The model identifies four occurrences of $f_2$ as shown as dark blue regions in Figure~\ref{fig: final derivations and library}. One might question why the model did not identify the two \texttt{B\%7 E7 Am7} successions as the \texttt{ii-V-I}. This is because if one uses $f_2$ for \texttt{B\%7 E7 Am7}, then it is summarized as a non-terminal \texttt{Am7} which makes the preceding \texttt{FM7} not interpretable (as the only way to incorporate it would be to see it as forming the \texttt{VI-V} predominant relation to \texttt{E7} which will break $f_2$). 

Patterns that generate both terminals and non-terminals are represented as rectangles with square holes (e.g., $f_4$ and $f_6$). The filled portion corresponds to surface chords that the pattern directly matches, while the hole marks an open argument that must be filled by another abstraction. Each such pattern has a fixed segment length (i.e., the number of terminal chords it consumes directly).

The learned abstractions of this type fall into two broad classes. 
The first consists of patterns formed by composing a primitive production with the termination rule, effectively introducing left- or right-branching structure to the primitive. Examples include $f_3$, a left-branching \texttt{V–I}; $f_4$, a left-branching descending fifth progression; and $f_5$, a left-branching backdoor dominant. These abstractions encode structurally incomplete constituents. For example, $f_3$ captures the notion of ``tonic missing a dominant on the left.'' In this respect, they resemble directional function application in categorical grammar \cite{steedman1993categorial}, where constituents specify how they combine with surrounding context. Interestingly, the model prioritizes left branching patterns. This observation is consistent with the goal directedness of previous grammar induction result \cite{harasim2020learnability}.
The second class comprises more structurally involved patterns that combine multiple non-terminating productions. Among these, $f_6$ is particularly noteworthy. It consumes two concrete surface chords surrounding a \texttt{bVII} harmonic region, where the initial chord stands in a descending-fifth relation to that region. The model surprisingly uses this abstraction to analyze the opening of \emph{Red Clay} in an unconventional way. Although the interpretation of the sequence \texttt{Bbm7-Dbsus-Ebsus} aligns with the expert analysis in the Jazz Harmony Tree Bank (JHTB) \cite{harasim2020jazz}, the treatment of the initial tonic differs. In the JHTB, the initial tonic is at the highest structural level. The model, however, interprets it as a descending-fifth of \texttt{Fsus} (the fifth chord). From the compression perspective, this interpretation makes sense because it can reuse the $f_6$ abstraction and draw an analogy between the two nested purple region \texttt{Cm7 \_ Fsus} and \texttt{Bbm7 \_ Ebsus} in a syntactically coherent way (see Figure \ref{fig: final derivations and library}).

Another interesting aspect is the usage of pattern $f_9$ in the piece \emph{Sunny} to explain the repeated segment \texttt{C7-FM7-B\%7-E7-Am7}. 
This is a typical \texttt{III-VI-ii-V-i} progression. In major mode, such roman numerals can be explained as a chain of applied dominant, where the \texttt{III} is the dominant of \texttt{VI}, which is the dominant of \texttt{ii}. However, upon closer inspection, such an interpretation is not valid in the minor mode context, where the root interval between \texttt{VI} and \texttt{ii} is a diminished fifth rather than a perfect fifth. This, together with the fact that the \texttt{VI} is of major seventh quality, eliminates the possibility to see the \texttt{VI} as a applied dominant of \texttt{II}. The model correctly interprets this five chords using \texttt{VI-V} relation on \texttt{FM7} and \texttt{E7}, which is consistent with the expert annotation in JHTB.

\section{General discussion \& Conclusion}

This paper proposed a new approach for finding jazz harmony patterns from observed surface sequence using library learning techniques, in which musical pattern discovery is formalized as the joint construction of derivations and reusable abstractions. At a conceptual level, the approach takes seriously a familiar metaphor in music theory: musical structure is built from reusable components. Here, this metaphor is made precise. Harmonic analyses are modeled as compositional programs constructed from primitive relations, and patterns are learned as reusable abstractions that can be composed systematically to generate complete derivations.
Musical patterns not as surface fragments, but as typed, compositional building blocks. They are not merely frequent substrings of chord symbols, nor simple concatenations of surface tokens. 

This raises a practical and conceptual question: how should musical patterns be encapsulated so that they can be used without recalling their full internal derivation? The answer lies in their type signature, which is the composed logical relations expressed as constraints. For example, we may summarize a learned abstraction such as 
$f_9$ without expanding it back into primitive rules as
\[
\mathrm{I} \;\to\; \mathrm{I}-(\mathrm{I})-\mathrm{III}^7-\mathrm{VI}-\mathrm{ii}-\mathrm{V}^7-\mathrm{I}
\]
What matters for its correct application is the set of constraints governing both terminals and non-terminals in this expansion. These constraints ensure that the pattern can be matched efficiently and that multiple chords can be generated or predicted in a single step. 
From the standpoint of a listener, parsing a piece can be likened to covering a sequence of chords with compatible structural blocks (as in Figure~\ref{fig: final derivations and library}) such that the overall structure remains coherent. If only small primitive relations are available, constructing a full analysis requires many incremental steps. Larger abstractions act as prefabricated blocks: once their surface and structural constraints are satisfied, they allow multiple chords to be incorporated in a single step, reducing search depth during inference. This also complements the limitation from PCFG's independence assumptions, by additionally specifying which combinations of production rules are likely to occur together. Cognitively, it suggests how internalized schemata might accelerate structural interpretation by allowing listeners to ``chunk'' sequences into higher-order units. Musical experience, to some extent, is captures by such library of musical abstractions one build over times.

Moreover, whether compression reliably produces musically plausible analyses remains an open empirical question. In the case of \emph{Red Clay}, the model achieves substantial compression, but the resulting derivation is unconventional: the opening tonic is assigned a deep structural role that departs from standard harmonic interpretations. This may reflect a limitation of the small corpus used in the proof-of-concept experiment. To address this limitation, future work can focus on improving the efficiency of implementation. With a larger and more diverse dataset, highly specific or niche analyses may no longer be compression-optimal, as patterns that generalize across many pieces would be favored. At the same time, conceiving music understanding as compression-guided library learning offers a new perspective on music theory itself. It raises the possibility that theoretical constructs such as harmonic schemata and formal archetypes emerge from implicit abstraction and reuse over repeated musical experience. 


\bibliographystyle{ACM-Reference-Format}
\bibliography{main}


\begin{thebibliography}{35}


\ifx \showCODEN    \undefined \def \showCODEN     #1{\unskip}     \fi
\ifx \showISBNx    \undefined \def \showISBNx     #1{\unskip}     \fi
\ifx \showISBNxiii \undefined \def \showISBNxiii  #1{\unskip}     \fi
\ifx \showISSN     \undefined \def \showISSN      #1{\unskip}     \fi
\ifx \showLCCN     \undefined \def \showLCCN      #1{\unskip}     \fi
\ifx \shownote     \undefined \def \shownote      #1{#1}          \fi
\ifx \showarticletitle \undefined \def \showarticletitle #1{#1}   \fi
\ifx \showURL      \undefined \def \showURL       {\relax}        \fi
\providecommand\bibfield[2]{#2}
\providecommand\bibinfo[2]{#2}
\providecommand\natexlab[1]{#1}
\providecommand\showeprint[2][]{arXiv:#2}

\bibitem[Boot et~al\mbox{.}(2016)]%
        {boot2016evaluating}
\bibfield{author}{\bibinfo{person}{Peter Boot}, \bibinfo{person}{Anja Volk}, {and} \bibinfo{person}{W~Bas de Haas}.} \bibinfo{year}{2016}\natexlab{}.
\newblock \showarticletitle{Evaluating the role of repeated patterns in folk song classification and compression}.
\newblock \bibinfo{journal}{\emph{Journal of New Music Research}} \bibinfo{volume}{45}, \bibinfo{number}{3} (\bibinfo{year}{2016}), \bibinfo{pages}{223--238}.
\newblock


\bibitem[Cambouropoulos(2006)]%
        {cambouropoulos2006musical}
\bibfield{author}{\bibinfo{person}{Emilios Cambouropoulos}.} \bibinfo{year}{2006}\natexlab{}.
\newblock \showarticletitle{Musical parallelism and melodic segmentation:: A computational approach}.
\newblock \bibinfo{journal}{\emph{Music Perception}} \bibinfo{volume}{23}, \bibinfo{number}{3} (\bibinfo{year}{2006}), \bibinfo{pages}{249--268}.
\newblock


\bibitem[Cao et~al\mbox{.}(2023)]%
        {cao2023babble}
\bibfield{author}{\bibinfo{person}{David Cao}, \bibinfo{person}{Rose Kunkel}, \bibinfo{person}{Chandrakana Nandi}, \bibinfo{person}{Max Willsey}, \bibinfo{person}{Zachary Tatlock}, {and} \bibinfo{person}{Nadia Polikarpova}.} \bibinfo{year}{2023}\natexlab{}.
\newblock \showarticletitle{babble: Learning better abstractions with e-graphs and anti-unification}.
\newblock \bibinfo{journal}{\emph{Proceedings of the ACM on Programming Languages}} \bibinfo{volume}{7}, \bibinfo{number}{POPL} (\bibinfo{year}{2023}), \bibinfo{pages}{396--424}.
\newblock


\bibitem[Cerna and Kutsia(2023)]%
        {cerna2023anti}
\bibfield{author}{\bibinfo{person}{David~M Cerna} {and} \bibinfo{person}{Temur Kutsia}.} \bibinfo{year}{2023}\natexlab{}.
\newblock \showarticletitle{Anti-unification and generalization: a survey}. In \bibinfo{booktitle}{\emph{Proceedings of the Thirty-Second International Joint Conference on Artificial Intelligence}}. \bibinfo{pages}{6563--6573}.
\newblock


\bibitem[Chakrabarty et~al\mbox{.}(2022)]%
        {chakrabarty2022approach}
\bibfield{author}{\bibinfo{person}{Sudipta Chakrabarty}, \bibinfo{person}{Ruhul Islam}, \bibinfo{person}{Emil Pricop}, {and} \bibinfo{person}{Hiren Kumar~Deva Sarma}.} \bibinfo{year}{2022}\natexlab{}.
\newblock \showarticletitle{An approach to discover similar musical patterns}.
\newblock \bibinfo{journal}{\emph{IEEE Access}}  \bibinfo{volume}{10} (\bibinfo{year}{2022}), \bibinfo{pages}{47322--47339}.
\newblock


\bibitem[Chater and Oaksford(2013)]%
        {chater2013programs}
\bibfield{author}{\bibinfo{person}{Nick Chater} {and} \bibinfo{person}{Mike Oaksford}.} \bibinfo{year}{2013}\natexlab{}.
\newblock \showarticletitle{Programs as causal models: Speculations on mental programs and mental representation}.
\newblock \bibinfo{journal}{\emph{Cognitive science}} \bibinfo{volume}{37}, \bibinfo{number}{6} (\bibinfo{year}{2013}), \bibinfo{pages}{1171--1191}.
\newblock


\bibitem[Cope(2002)]%
        {cope2002recombinant}
\bibfield{author}{\bibinfo{person}{David Cope}.} \bibinfo{year}{2002}\natexlab{}.
\newblock \showarticletitle{Recombinant music: using the computer to explore musical style}.
\newblock \bibinfo{journal}{\emph{Computer}} \bibinfo{volume}{24}, \bibinfo{number}{7} (\bibinfo{year}{2002}), \bibinfo{pages}{22--28}.
\newblock


\bibitem[de~Haas et~al\mbox{.}(2013)]%
        {de2013structural}
\bibfield{author}{\bibinfo{person}{W~Bas de Haas}, \bibinfo{person}{Anja Volk}, {and} \bibinfo{person}{Frans Wiering}.} \bibinfo{year}{2013}\natexlab{}.
\newblock \showarticletitle{Structural segmentation of music based on repeated harmonies}. In \bibinfo{booktitle}{\emph{2013 IEEE International Symposium on Multimedia}}. IEEE, \bibinfo{pages}{255--258}.
\newblock


\bibitem[De~Leon and Inesta(2007)]%
        {de2007pattern}
\bibfield{author}{\bibinfo{person}{Pedro J~Ponce De~Leon} {and} \bibinfo{person}{Jos~M Inesta}.} \bibinfo{year}{2007}\natexlab{}.
\newblock \showarticletitle{Pattern recognition approach for music style identification using shallow statistical descriptors}.
\newblock \bibinfo{journal}{\emph{IEEE Transactions on Systems, Man, and Cybernetics, Part C (Applications and Reviews)}} \bibinfo{volume}{37}, \bibinfo{number}{2} (\bibinfo{year}{2007}), \bibinfo{pages}{248--257}.
\newblock


\bibitem[Hall and Pearce(2021)]%
        {hall2021model}
\bibfield{author}{\bibinfo{person}{Edward~TR Hall} {and} \bibinfo{person}{Marcus~T Pearce}.} \bibinfo{year}{2021}\natexlab{}.
\newblock \showarticletitle{A model of large-scale thematic structure}.
\newblock \bibinfo{journal}{\emph{Journal of New Music Research}} \bibinfo{volume}{50}, \bibinfo{number}{3} (\bibinfo{year}{2021}), \bibinfo{pages}{220--241}.
\newblock


\bibitem[Harasim(2020)]%
        {harasim2020learnability}
\bibfield{author}{\bibinfo{person}{Daniel Harasim}.} \bibinfo{year}{2020}\natexlab{}.
\newblock \emph{\bibinfo{title}{The learnability of the grammar of jazz: Bayesian inference of hierarchical structures in harmony}}.
\newblock \bibinfo{thesistype}{Ph.\,D. Dissertation}. \bibinfo{school}{EPFL}.
\newblock


\bibitem[Harasim et~al\mbox{.}(2020)]%
        {harasim2020jazz}
\bibfield{author}{\bibinfo{person}{Daniel Harasim}, \bibinfo{person}{Christoph Finkensiep}, \bibinfo{person}{Petter Ericson}, \bibinfo{person}{Timothy~J O’Donnell}, {and} \bibinfo{person}{Martin Rohrmeier}.} \bibinfo{year}{2020}\natexlab{}.
\newblock \showarticletitle{The jazz harmony treebank}. In \bibinfo{booktitle}{\emph{21st ISMIR, Montr{\'e}al, Canada, October 11-16, 2020}}. \bibinfo{pages}{207--215}.
\newblock


\bibitem[Hsiao et~al\mbox{.}(2023)]%
        {hsiao2023bps}
\bibfield{author}{\bibinfo{person}{Yo-Wei Hsiao}, \bibinfo{person}{Tzu-Yun Hung}, \bibinfo{person}{Tsung-Ping Chen}, {and} \bibinfo{person}{Li Su}.} \bibinfo{year}{2023}\natexlab{}.
\newblock \showarticletitle{BPS-Motif: A Dataset for Repeated Pattern Discovery of Polyphonic Symbolic Music.}. In \bibinfo{booktitle}{\emph{ISMIR}}. \bibinfo{pages}{281--288}.
\newblock


\bibitem[Janssen et~al\mbox{.}(2013)]%
        {janssen2013finding}
\bibfield{author}{\bibinfo{person}{Berit Janssen}, \bibinfo{person}{W~Bas De~Haas}, \bibinfo{person}{Anja Volk}, {and} \bibinfo{person}{Peter Van~Kranenburg}.} \bibinfo{year}{2013}\natexlab{}.
\newblock \showarticletitle{Finding repeated patterns in music: State of knowledge, challenges, perspectives}. In \bibinfo{booktitle}{\emph{International Symposium on Computer Music Multidisciplinary Research}}. Springer, \bibinfo{pages}{277--297}.
\newblock


\bibitem[Lake et~al\mbox{.}(2015)]%
        {lake2015human}
\bibfield{author}{\bibinfo{person}{Brenden~M Lake}, \bibinfo{person}{Ruslan Salakhutdinov}, {and} \bibinfo{person}{Joshua~B Tenenbaum}.} \bibinfo{year}{2015}\natexlab{}.
\newblock \showarticletitle{Human-level concept learning through probabilistic program induction}.
\newblock \bibinfo{journal}{\emph{Science}} \bibinfo{volume}{350}, \bibinfo{number}{6266} (\bibinfo{year}{2015}), \bibinfo{pages}{1332--1338}.
\newblock


\bibitem[Lartillot and Toiviainen(2007)]%
        {lartillot2007motivic}
\bibfield{author}{\bibinfo{person}{Olivier Lartillot} {and} \bibinfo{person}{Petri Toiviainen}.} \bibinfo{year}{2007}\natexlab{}.
\newblock \showarticletitle{Motivic matching strategies for automated pattern extraction}.
\newblock \bibinfo{journal}{\emph{Musicae Scientiae}} \bibinfo{volume}{11}, \bibinfo{number}{1\_suppl} (\bibinfo{year}{2007}), \bibinfo{pages}{281--314}.
\newblock


\bibitem[Nieto et~al\mbox{.}(2020)]%
        {nieto2020audio}
\bibfield{author}{\bibinfo{person}{Oriol Nieto}, \bibinfo{person}{Gautham~J Mysore}, \bibinfo{person}{Cheng-i Wang}, \bibinfo{person}{Jordan~BL Smith}, \bibinfo{person}{Jan Schl{\"u}ter}, \bibinfo{person}{Thomas Grill}, {and} \bibinfo{person}{Brian McFee}.} \bibinfo{year}{2020}\natexlab{}.
\newblock \showarticletitle{Audio-based music structure analysis: Current trends, open challenges, and applications}.
\newblock \bibinfo{journal}{\emph{Transactions of the International Society for Music Information Retrieval}} \bibinfo{volume}{3}, \bibinfo{number}{1} (\bibinfo{year}{2020}).
\newblock


\bibitem[Paulus and Klapuri(2006)]%
        {paulus2006music}
\bibfield{author}{\bibinfo{person}{Jouni Paulus} {and} \bibinfo{person}{Anssi Klapuri}.} \bibinfo{year}{2006}\natexlab{}.
\newblock \showarticletitle{Music structure analysis by finding repeated parts}. In \bibinfo{booktitle}{\emph{Proceedings of the 1st ACM workshop on Audio and music computing multimedia}}. \bibinfo{pages}{59--68}.
\newblock


\bibitem[Paulus et~al\mbox{.}(2010)]%
        {paulus2010state}
\bibfield{author}{\bibinfo{person}{Jouni Paulus}, \bibinfo{person}{Meinard M{\"u}ller}, {and} \bibinfo{person}{Anssi Klapuri}.} \bibinfo{year}{2010}\natexlab{}.
\newblock \showarticletitle{State of the Art Report: Audio-Based Music Structure Analysis.}. In \bibinfo{booktitle}{\emph{Ismir}}. Utrecht, \bibinfo{pages}{625--636}.
\newblock


\bibitem[Pearce and M{\"u}llensiefen(2017)]%
        {pearce2017compression}
\bibfield{author}{\bibinfo{person}{Marcus Pearce} {and} \bibinfo{person}{Daniel M{\"u}llensiefen}.} \bibinfo{year}{2017}\natexlab{}.
\newblock \showarticletitle{Compression-based modelling of musical similarity perception}.
\newblock \bibinfo{journal}{\emph{Journal of New Music Research}} \bibinfo{volume}{46}, \bibinfo{number}{2} (\bibinfo{year}{2017}), \bibinfo{pages}{135--155}.
\newblock


\bibitem[Pereira and Warren(1983)]%
        {pereira1983parsing}
\bibfield{author}{\bibinfo{person}{Fernando~CN Pereira} {and} \bibinfo{person}{David~HD Warren}.} \bibinfo{year}{1983}\natexlab{}.
\newblock \showarticletitle{Parsing as deduction}. In \bibinfo{booktitle}{\emph{21st annual meeting of the association for computational linguistics}}. \bibinfo{pages}{137--144}.
\newblock


\bibitem[Plotkin(1970)]%
        {plotkin1970lattice}
\bibfield{author}{\bibinfo{person}{Gordon Plotkin}.} \bibinfo{year}{1970}\natexlab{}.
\newblock \bibinfo{booktitle}{\emph{Lattice theoretic properties of subsumption}}.
\newblock \bibinfo{publisher}{Edinburgh University, Department of Machine Intelligence and Perception}.
\newblock


\bibitem[Rohrmeier(2020)]%
        {rohrmeier2020syntax}
\bibfield{author}{\bibinfo{person}{Martin Rohrmeier}.} \bibinfo{year}{2020}\natexlab{}.
\newblock \showarticletitle{The syntax of jazz harmony: Diatonic tonality, phrase structure, and form}.
\newblock \bibinfo{journal}{\emph{Music Theory and Analysis (MTA)}} \bibinfo{volume}{7}, \bibinfo{number}{1} (\bibinfo{year}{2020}), \bibinfo{pages}{1--63}.
\newblock


\bibitem[Ross et~al\mbox{.}(2012)]%
        {ross2012detecting}
\bibfield{author}{\bibinfo{person}{Joe~Cheri Ross}, \bibinfo{person}{TP Vinutha}, {and} \bibinfo{person}{Preeti Rao}.} \bibinfo{year}{2012}\natexlab{}.
\newblock \showarticletitle{Detecting Melodic Motifs from Audio for Hindustani Classical Music.}. In \bibinfo{booktitle}{\emph{ISMIR}}. \bibinfo{pages}{193--198}.
\newblock


\bibitem[Sargent et~al\mbox{.}(2011)]%
        {sargent2011regularity}
\bibfield{author}{\bibinfo{person}{Gabriel Sargent}, \bibinfo{person}{Fr{\'e}d{\'e}ric Bimbot}, {and} \bibinfo{person}{Emmanuel Vincent}.} \bibinfo{year}{2011}\natexlab{}.
\newblock \showarticletitle{A regularity-constrained Viterbi algorithm and its application to the structural segmentation of songs}. In \bibinfo{booktitle}{\emph{International Society for Music Information Retrieval Conference (ISMIR)}}.
\newblock


\bibitem[Shieber et~al\mbox{.}(1995)]%
        {shieber1995principles}
\bibfield{author}{\bibinfo{person}{Stuart~M Shieber}, \bibinfo{person}{Yves Schabes}, {and} \bibinfo{person}{Fernando~CN Pereira}.} \bibinfo{year}{1995}\natexlab{}.
\newblock \showarticletitle{Principles and implementation of deductive parsing}.
\newblock \bibinfo{journal}{\emph{The Journal of logic programming}} \bibinfo{volume}{24}, \bibinfo{number}{1-2} (\bibinfo{year}{1995}), \bibinfo{pages}{3--36}.
\newblock


\bibitem[Steedman(1993)]%
        {steedman1993categorial}
\bibfield{author}{\bibinfo{person}{Mark Steedman}.} \bibinfo{year}{1993}\natexlab{}.
\newblock \showarticletitle{Categorial grammar}.
\newblock \bibinfo{journal}{\emph{Lingua}} \bibinfo{volume}{90}, \bibinfo{number}{3} (\bibinfo{year}{1993}), \bibinfo{pages}{221--258}.
\newblock


\bibitem[Tate et~al\mbox{.}(2009)]%
        {tate2009equality}
\bibfield{author}{\bibinfo{person}{Ross Tate}, \bibinfo{person}{Michael Stepp}, \bibinfo{person}{Zachary Tatlock}, {and} \bibinfo{person}{Sorin Lerner}.} \bibinfo{year}{2009}\natexlab{}.
\newblock \showarticletitle{Equality saturation: a new approach to optimization}. In \bibinfo{booktitle}{\emph{Proceedings of the 36th annual ACM SIGPLAN-SIGACT symposium on Principles of programming languages}}. \bibinfo{pages}{264--276}.
\newblock


\bibitem[Tenenbaum et~al\mbox{.}(2011)]%
        {tenenbaum2011grow}
\bibfield{author}{\bibinfo{person}{Joshua~B Tenenbaum}, \bibinfo{person}{Charles Kemp}, \bibinfo{person}{Thomas~L Griffiths}, {and} \bibinfo{person}{Noah~D Goodman}.} \bibinfo{year}{2011}\natexlab{}.
\newblock \showarticletitle{How to grow a mind: Statistics, structure, and abstraction}.
\newblock \bibinfo{journal}{\emph{science}} \bibinfo{volume}{331}, \bibinfo{number}{6022} (\bibinfo{year}{2011}), \bibinfo{pages}{1279--1285}.
\newblock


\bibitem[Van~Kranenburg and Backer(2005)]%
        {van2005musical}
\bibfield{author}{\bibinfo{person}{Peter Van~Kranenburg} {and} \bibinfo{person}{Eric Backer}.} \bibinfo{year}{2005}\natexlab{}.
\newblock \showarticletitle{Musical style recognition—a quantitative approach}.
\newblock In \bibinfo{booktitle}{\emph{Handbook of pattern recognition and computer vision}}. \bibinfo{publisher}{World Scientific}, \bibinfo{pages}{583--600}.
\newblock


\bibitem[Volk and Van~Kranenburg(2012)]%
        {volk2012melodic}
\bibfield{author}{\bibinfo{person}{Anja Volk} {and} \bibinfo{person}{Peter Van~Kranenburg}.} \bibinfo{year}{2012}\natexlab{}.
\newblock \showarticletitle{Melodic similarity among folk songs: An annotation study on similarity-based categorization in music}.
\newblock \bibinfo{journal}{\emph{Musicae Scientiae}} \bibinfo{volume}{16}, \bibinfo{number}{3} (\bibinfo{year}{2012}), \bibinfo{pages}{317--339}.
\newblock


\bibitem[Willsey et~al\mbox{.}(2021)]%
        {willsey2021egg}
\bibfield{author}{\bibinfo{person}{Max Willsey}, \bibinfo{person}{Chandrakana Nandi}, \bibinfo{person}{Yisu~Remy Wang}, \bibinfo{person}{Oliver Flatt}, \bibinfo{person}{Zachary Tatlock}, {and} \bibinfo{person}{Pavel Panchekha}.} \bibinfo{year}{2021}\natexlab{}.
\newblock \showarticletitle{Egg: Fast and extensible equality saturation}.
\newblock \bibinfo{journal}{\emph{Proceedings of the ACM on Programming Languages}} \bibinfo{volume}{5}, \bibinfo{number}{POPL} (\bibinfo{year}{2021}), \bibinfo{pages}{1--29}.
\newblock


\bibitem[Zhang et~al\mbox{.}(2023)]%
        {zhang2023better}
\bibfield{author}{\bibinfo{person}{Yihong Zhang}, \bibinfo{person}{Yisu~Remy Wang}, \bibinfo{person}{Oliver Flatt}, \bibinfo{person}{David Cao}, \bibinfo{person}{Philip Zucker}, \bibinfo{person}{Eli Rosenthal}, \bibinfo{person}{Zachary Tatlock}, {and} \bibinfo{person}{Max Willsey}.} \bibinfo{year}{2023}\natexlab{}.
\newblock \showarticletitle{Better together: Unifying datalog and equality saturation}.
\newblock \bibinfo{journal}{\emph{Proceedings of the ACM on Programming Languages}} \bibinfo{volume}{7}, \bibinfo{number}{PLDI} (\bibinfo{year}{2023}), \bibinfo{pages}{468--492}.
\newblock


\bibitem[Zhang et~al\mbox{.}(2022)]%
        {zhang2022influence}
\bibfield{author}{\bibinfo{person}{Yu Zhang}, \bibinfo{person}{Ziya Zhou}, {and} \bibinfo{person}{Maosong Sun}.} \bibinfo{year}{2022}\natexlab{}.
\newblock \showarticletitle{Influence of musical elements on the perception of ‘Chinese style’in music}.
\newblock \bibinfo{journal}{\emph{Cognitive Computation and Systems}} \bibinfo{volume}{4}, \bibinfo{number}{2} (\bibinfo{year}{2022}), \bibinfo{pages}{147--164}.
\newblock


\bibitem[Ziv and Eitan(2007)]%
        {ziv2007themes}
\bibfield{author}{\bibinfo{person}{Naomi Ziv} {and} \bibinfo{person}{Zohar Eitan}.} \bibinfo{year}{2007}\natexlab{}.
\newblock \showarticletitle{Themes as prototypes: Similarity judgments and categorization tasks in musical contexts}.
\newblock \bibinfo{journal}{\emph{Musicae Scientiae}} \bibinfo{volume}{11}, \bibinfo{number}{1\_suppl} (\bibinfo{year}{2007}), \bibinfo{pages}{99--133}.
\newblock


\end{thebibliography}

\pagebreak
\appendix

\section{Encodings in \egglog}
\begin{listing}[h]
\begin{minted}{lisp}
;; This rule essentially says if 
;; 1) x's root is P5 above the y's root, 
;; 2) x's core quality is not dominant seventh, 
;; then x y forms a descending fifth relation
;; and the pair can be reduced to y. 
(rule
    ( (= x (ChordLabel rX qX _))
      (= y (ChordLabel rY qY _))
      (IntervalDown rX rY "P5")
      (!= qX (StackOfThirds (Maj) (Min) (Min))) )
    ( (ExpandsInto y (Prep (Descending5th)) x y) )
    :ruleset grammar-rules
)
    \end{minted}
\caption{Encoding the \texttt{Descending5th} rule in \egglog. The rules consider chord qualities as equivalent up to the essential seventh-chord structure. 
For example, major seventh = (M3,m3,M3); dominant seventh = (M3,m3,m3).
}
\label{fig: Encoding the Descending5th rule in egglog}
\end{listing}

\begin{listing}[h]
\begin{minted}{lisp}
(ruleset cyk-rules)
(rule
    ( (IsWord title t i)       
      (TerminatesAs nt r t) )
    ( (IsPhrase title nt i (+ i 1))          
      (union                          
        (Pure r)                   
        (TemplateQueryOnPhrase title nt i (+ i 1))))
    :ruleset cyk-rules)
(rule
    ( (IsPhrase title y i j)    
      (IsPhrase title z j k)    
      (ExpandsInto x r y z) )
    ( (IsPhrase title x i k) 
      (union 
        (WithRep (Pure r) [New,New] 
            [ (TemplateQueryOnPhrase title y i j)
            , (TemplateQueryOnPhrase title z j k)
            ])
        (TemplateQueryOnPhrase title x i k) ) )
    :ruleset cyk-rules)
\end{minted}
\caption{Adapting CYK-parsing rules in \egglog to explicitly store all the template program derivations in the e-graph. We uses bracket list notation instead of the traditional \texttt{Cons}/ \texttt{Nil} constructors to make the code more readable.}
\end{listing}

\newpage
\onecolumn
\section{Inference rules}

\begin{figure*}[h!]
  \centering
  \input{rules/cyk-rules}
  \caption{Representative \egglog inference rules for the parsing stage}
  \label{fig: cyk_rules}
\end{figure*}

\begin{figure*}[h!]
  \centering
  \input{rules/Au-Rules}
  \caption{Anti-unification rules. Antiunifiers are set to be finalized after these rules saturates to fixed point. This cycle between deducing antiunifiers and finalizing antiunifiers also runs to a fixed point.}
\end{figure*}

\begin{figure*}[h!]
  \centering
  \input{rules/cost-set-analysis-rules}
  \caption{\babble's Cost set analysis implemented via inference rules. Prune and Reduce are two  set filtering function from \babble. Reduce eliminates provably worse pairs according to a partial order, and Prune is an auxiliary function for beam search to keep only the top k candidate according to use cost. NodeNeeded is a function that precalculate e-nodes that requires analysis value in a e-class, and NodeProcessed is growing counter (represent by a set) to keep track of the collection process. }
\end{figure*}

\end{document}